# Active Learning of Halfspaces under a Margin Assumption

**Alon Gonen**                                    ALONGNN@CS.HUJI.AC.IL
**Sivan Sabato**                                  SABATO@CS.HUJI.AC.IL
**Shai Shalev-Shwartz**                           SHAIS@CS.HUJI.AC.IL
*Benin school of Computer Science and Engineering*
*The Hebrew University*
*Givat Ram, Jerusalem 91904, Israel*

## Abstract

We derive and analyze a new, efficient, pool-based active learning algorithm for halfspaces, called ALuMA. Most previous algorithms show exponential improvement in the label complexity assuming that the distribution over the instance space is close to uniform. This assumption rarely holds in practical applications. Instead, we study the label complexity under a large-margin assumption—a much more realistic condition, as evident by the success of margin-based algorithms such as SVM. Our algorithm is computationally efficient and comes with formal guarantees on its label complexity. It also naturally extends to the non-separable case and to non-linear kernels. Experiments illustrate the clear advantage of ALuMA over other active learning algorithms.

## 1. Introduction

We consider the challenge of pool-based active learning (McCallum and Nigam, 1998), in which a learner receives a pool of unlabeled examples, and can iteratively query a teacher for the labels of examples from the pool. The goal of the learner is to return a low-error prediction rule using a small number of queries. The number of queries used by the learner is termed its *label complexity*. This setting is most useful when unlabeled data is abundant but labeling is expensive, a common case in many data-laden applications.

In some cases, pool-based active learning can provide an exponential improvement in label complexity over standard 'passive' learning. For instance, suppose the examples are points in $[0, 1]$ and there exists some unknown threshold such that points below the threshold are classified as negative and points above the threshold are classified as positive. The sample complexity of learning a prediction rule with error less than $\epsilon$ using an i.i.d. labeled sample is $\tilde{\Theta}(1/\epsilon)$. In comparison, an active learner can select its queries to follow a binary search, and thus can achieve the same accuracy with a label complexity of only $O(\ln(1/\epsilon))$. This result holds for any distribution over $[0, 1]$.

The example above is a special and very simple case of the important and highly common setting of active learning of halfspaces in $\mathbb{R}^d$. Much research has been devoted to this challenge, and several algorithms have been proposed. For instance, the Query By Committee (QBC) algorithm (Seung et al., 1992; Freund et al., 1997) and the CAL algorithm (Cohn et al., 1994) both examine the unlabeled examples sequentially, and maintain a version space that is shrunk with each received label. QBC requests the label of an example if two hypotheses randomly sampled from the current version space disagree on its label. CAL simply requests the label of any example whose label is not determined by the current version space. Another example is an active variant of the Perceptron algorithm, proposed in Dasgupta et al. (2005).

If the marginal distribution of the data is uniform over a sphere centered at the origin, then all of these algorithms achieve an exponential improvement in the label complexity compared to passive learning, similarly to the single-dimensional case of thresholds on the line.

For CAL, a more general result has been shown: a label complexity of $O(\ln(1/\epsilon))$ is achieved whenever the joint distribution of data and labels has a bounded disagreement coefficient (Hanneke, 2007, 2009). This holds, for instance, for any "smooth" distribution (Friedman, 2009). However, the "O notation" here hides a dependency on the disagreement coefficient, which for some data distributions can be very large, making this upper bound similar to the sample complexity required by a passive learner. This makes the exponential improvement with respect to $\epsilon$ meaningless in these cases.

To illustrate this point, we consider the following example, due to Dasgupta (2006), showing a simple distribution in $\mathbb{R}^3$ for which no significant improvement over passive learning can be achieved with any active learning algorithm.

**Example 1** *Consider a distribution in $\mathbb{R}^d$ for any $d \geq 3$. Suppose that the support of the distribution is a set of evenly-distributed points on a two-dimensional sphere that does not circumscribe the origin, as illustrated in the following figure. As can be seen, each point can be separated from the rest of the points with a halfspace.*

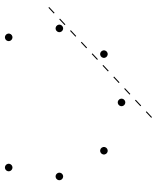

In this example, to distinguish between the case in which all points have a negative label and the case in which one of the points has a positive label while the rest have a negative label, any active learning algorithm will have to query every point at least once. It follows that for any $\epsilon > 0$, if the number of points is $1/\epsilon$, then the label complexity to achieve an error of at most $\epsilon$ is $1/\epsilon$. On the other hand, the sample complexity of passive learning in this case is order of $\frac{1}{\epsilon} \log \frac{1}{\epsilon}$, hence no active learner can be significantly better than a passive learner on this distribution.[1] This proves the following claim:

**Claim 1** *Let $C$ be a universal constant. For all $\epsilon \in (0, \frac{1}{2})$ and $d \geq 3$, there exists a distribution over $\mathbb{R}^d$, such that no active learner can have label complexity smaller than $1/\epsilon$ for learning the hypothesis class of origin-centered halfspaces on this distribution. On the* other hand, the sample complexity of the vanilla ERM passive learner for this distribution is $\frac{C}{\epsilon} \log \frac{1}{\epsilon}$.

---

1. The disagreement coefficient for this example is equal to the number of points. In addition, while the distribution in this example is not smooth, and thus the results of Friedman (2009) do not hold here, we can easily slightly change the distribution to make it smooth, while still having the same phenomenon. This does not contradict the result of Friedman (2009), as the exponential improvement from $1/\epsilon$ to $\ln(1/\epsilon)$ kicks in only when $1/\epsilon$ is larger than the number of points.

We see that there cannot be a true exponential improvement with respect to $\epsilon$ which is uniform over all distributions and target hypotheses. Moreover, the only known case where a non-trivial label complexity bound can be achieved for active learning is when the distribution is uniform (or nearly uniform) over a sphere centered at the origin. This is a serious limitation, since real applications rarely exhibit anything similar to a uniform distribution of their data.

A second limitation of the result for CAL is that despite the logarithmic dependence on $1/\epsilon$, the absolute label complexity of CAL can be much worse than that of the optimal algorithm. This is illustrated in the following example and the theorem following it.

**Example 2** *Consider a distribution in $\mathbb{R}^d$ that is supported by two types of points on an octahedron (see an illustration for $\mathbb{R}^3$ below).*

1. *Vertices: $\{e_1, \ldots, e_d\}$.*

2. *Face centers: $z/d$ for $z \in \{-1, +1\}^d$.*

*Consider the hypothesis class $\mathcal{W} = \{x \mapsto \operatorname{sgn}(\langle x, w\rangle - 1 + \frac{1}{d}) \mid w \in \{-1, +1\}^d\}$. Each hypothesis in $\mathcal{W}$, defined by some $w \in \{-1, +1\}^d$, classifies at most $d + 1$ data points as positive: these are the vertices $e_i$ for $i$ such that $w[i] = +1$, and the face center $w/d$.*

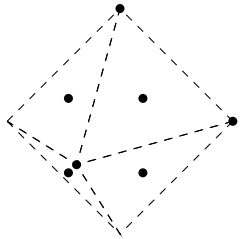

**Theorem 1** *Consider Example 2 for $d \geq 3$, and assume that the pool of examples includes the entire support of the distribution. There is an efficient algorithm that finds the correct hypothesis from $\mathcal{W}$ with at most $d$ labels. On the other hand, with probability at least $\frac{1}{e}$ over the randomization of the sample, CAL uses at least $\frac{2^d + d}{2d + 3}$ labels to find the correct separator.*

**Proof** First, it is easy to see that if $h^* \in \mathcal{W}$ is the correct hypothesis, then

$$w = (h^*(e_1), \ldots, h^*(e_d)).$$

Thus, it suffices to query the $d$ vertices to discover the true $w$.

We now show that the number of queries CAL asks until finding the correct separator is exponential in $d$. Consider some run of CAL (determined by the random ordering of the sample). Assume w.l.o.g. that each data point appears once in the sample. Let $S$ be the set that includes the positive face center and all the vertices. Note that CAL cannot terminate before either querying all the $2^d - 1$ negative face centers, or querying at least one example from $S$. Moreover, CAL will query all the face centers it encounters before encountering the first example from $S$. At each iteration $t$ before encountering an example from $S$, there is a probability of $\frac{d+1}{2^d+d-t}$ that the next example is from $S$. Therefore, the probability that the first $T = \frac{2^d+d}{2d+3}$ examples are not from $S$ is

$$\prod_{t=0}^{T-1}\left(1 - \frac{d+1}{2^d+d-t}\right) \geq \left(1 - \frac{d+1}{2^d+d-T}\right)^T \geq e^{-2T\frac{d+1}{2^d+d-T}} = e^{\frac{-2(d+1)}{\frac{2^d+d}{T}-1}} = \frac{1}{e} \;,$$

where in the second equality we used $1 - a \geq \exp(-2a)$ which holds for all $a \in [0, \frac{1}{2}]$. Therefore, with probability at least $\frac{1}{e}$ the number of queries is at least $\frac{2^d+d}{2d+3}$. ∎

In this paper we present a new efficient pool-based active learning algorithm for learning high dimensional halfspaces. As Example 1 above shows, it is not possible to guarantee a significant improvement in label complexity that holds uniformly for all distributions. However, for any given pool of unlabeled examples, there exists *some* optimal label complexity—the one that would be achieved by an optimal active learning algorithm for this pool. By "optimal" we mean an algorithm which queries the minimal number of labels in the worst-case sense, where worst-case here is with respect to the target hypothesis. Unfortunately, it is unknown whether there exists a polynomial-time optimal algorithm for the class of halfspaces. We therefore present an efficient approximation algorithm. That is, we will analyze the label complexity of our algorithm with a *regret-type* analysis: we bound the label complexity of our algorithm relative to the optimal label complexity that can be achieved for the given unlabeled pool. Therefore, if for a given pool of examples active learning can outperform passive learning, then our algorithm will enjoy a similar improvement. On the other hand, if the pool is inherently "difficult", as in Example 1, our algorithm might require the same number of labels as a passive learner.

As in QBC and CAL, we maintain a version space and update it whenever we receive a label. Unlike QBC and CAL, we do not restrict ourselves to examining the unlabeled examples sequentially. Instead, we examine the entire pool at each iteration and select the next example to query in a greedy manner—we select an example that approximately maximizes the worst-case reduction in the version space.[2]

Deriving active learning algorithms by greedily maximizing reductions in the version space has been recently proposed by Golovin and Krause (2010). In particular, they proposed the notion of adaptive sub-modular optimization, and showed that in the case of a finite hypothesis class, a greedy selection rule can be competitive with the optimal algorithm, where the competitiveness factor depends on the logarithm of the size of the hypothesis class. However, in the case of halfspaces the hypothesis class is of infinite size, hence this analysis is not directly applicable.[3] Furthermore, a generic implementation of their algorithm yields a runtime that depends linearly on the size of the hypothesis class.

To tackle these issues, we will rely on the familiar notion of *margin*. We will assume that the negative and positive examples can be separated (or approximately separated) with a positive margin. Under this assumption, we are able to derive an *efficient* algorithm with a label-complexity competitiveness guarantee that depends on the margin. A larger margin will result in better regret guarantees for our algorithm. Some of the most popular learning algorithms (e.g. SVM, Perceptron, and AdaBoost) rely on a large margin assumption, and their practical success indicates that margin is a reasonable requirement for many real world problems.

---

2. The efficient algorithm shown in the proof of Theorem 1 is in fact an instance of this strategy.

3. By Sauer's lemma, the effective size of the hypothesis class of halfspaces on a set of $m$ unlabeled examples is only $m^d$. One can thus apply the algorithm of Golovin and Krause (2010) to this finite hypothesis class, however the runtime of the resulting algorithm will be exponential in $d$.

The margin assumption is beneficial in two additional aspects. First, our algorithm can handle high-dimensional data by performing a pre-processing step of dimensionality reduction. In particular, this can be performed when the data is represented by a kernel matrix. Kernels have had tremendous impact on machine learning theory and algorithms over the past decade (Cristianini and Shawe-Taylor, 2004; Schölkopf and Smola, 2002), so the ability to apply our algorithm with kernels is a clear advantage. We note that an efficient implementation of QBC with kernels has been proposed in Gilad-Bachrach et al. (2005), but no formal label complexity guarantees have been provided for the resulting algorithm.

Second, our algorithm can handle samples with a small label error via a simple transformation, that transforms any labeled sample to a sample which is separable with a margin. This margin depends on the optimal hinge-loss of the sample. This transformation can be applied for any algorithm, including CAL or QBC. However, the transformation leads to a data distribution which might be far from uniform, hence the label complexity guarantees of these algorithms do not apply. In contrast, since our analysis only assumes margin, the theoretical guarantees still hold for our algorithm.

We note that several algorithms have been proposed for active learning in the agnostic case, for instance $A^2$ (Balcan et al., 2006a) and IWAL (Beygelzimer et al., 2009). These algorithms provide similar exponential speedups for the uniform distribution without requiring separability. However, unlike the algorithms for the separable case, these algorithms are computationally intractable when learning the class of halfspaces with the zero-one loss.

To summarize, the proposed algorithm is computationally efficient, tolerant to some label error, applicable for learning in high dimensional spaces, and comes equipped with label complexity guarantees that hold for any distribution in $\mathbb{R}^d$ that (approximately) satisfies a margin assumption.

Tong and Koller (2002) also addressed active learning of halfspaces under the assumption of separability with a margin. They proposed the principle of choosing the example that splits the version space as evenly as possible. They show that if at every round the chosen example splits the version space exactly in half, then the algorithm achieves the minimal possible label complexity (in a certain Bayesian sense). They implement several heuristics that attempt to follow this selection principle using an efficient algorithm. For instance, they suggest to choose the vector $x$ which is closest to the max-margin solution of the data labeled so far. However, no formal guarantees are provided for these heuristics relative to the proposed principle. Moreover, for each of these heuristics there are cases where the split induced by the example selected by the heuristic is much less balanced than the one induced by the most balanced example. In Balcan et al. (2007), an active learning algorithm with guarantees under a margin assumption is proposed. However, these guarantees hold only if the data distribution is uniform over an ellipsoid.

One may wonder if a large margin assumption alone can guarantee a uniform improvement in label complexity. This is not the case, as evident by the following example.

**Example 3** *Let $\gamma \in (0, \frac{1}{2})$ be a margin parameter. Consider the uniform distribution supported by $N$ points in $\mathbb{R}^d$, such that all the points are on the unit sphere, and for each pair of points $x_1$ and $x_2$ in the support, $\langle x_1, x_2 \rangle \leq 1 - 2\gamma$. It was shown in Shannon (1959) that for any $N \leq O((1/\gamma)^d)$, there exists a set of points that satisfy the conditions above. Now, for any point $x$ in the support, there exists a (biased) linear separator that separates $x$ from the rest of the points with a margin of $\gamma$. This can be seen by letting $w = x$ and $b = 1 - \gamma$. Then $\langle w, x \rangle - b = \gamma$ while for any $z \neq x$ in the set, $\langle w, z \rangle - b = \langle x, z \rangle - 1 + \gamma \leq -\gamma$. By adding a single dimension, this example can be transformed to one with origin-centered halfspaces.*

Just as in Example 1, here too any active learner must query all $N$ points in the worst case. We have thus proved:

**Claim 2** *Let $C$ be a universal constant. For all $d \geq 2$, $\gamma \in (0, \frac{1}{2})$, $N = \Omega((1/\gamma)^{d-1})$ and $\epsilon \in (\frac{1}{N}, \frac{1}{2})$, there exists a distribution over $\mathbb{R}^d$, such that no active learning algorithm can have label complexity smaller than $1/\epsilon$ for learning the hypothesis class of origin-centered halfspaces on this distribution, even if the learner knows that the data is separable with margin of $\gamma$. On the other hand, the sample complexity of the vanilla ERM passive learner for this distribution is $\frac{C}{\epsilon} \ln(\frac{1}{\epsilon})$.*

We provide a formal problem statement in Section 2, and state our main results in Section 3. Section 4 describes the ALuMA algorithm, and Section 5 outlines the proof of the properties of ALuMA. Section 6 provides the transformation that allows ALuMA to handle kernel representations and labeling errors. In Section 7 we describe a simpler implementation of ALuMA, which is useful for running the algorithm in practice. We report experiments on real and synthetic datasets in Section 8. The detailed proofs of our main results are disclosed in Appendix A and Appendix B.

## 2. Formal Problem Statement

In pool-based active learning, the learner receives as input a set of instances, denoted $X = \{x_1, \ldots, x_m\}$. Each instance is associated with a label (which is initially unknown to the learner). The goal of the active learner is to label all $x \in X$ correctly using as few label queries as possible. If $X$ is a sample drawn i.i.d. from a fixed distribution, then the labels the active learning algorithm outputs can then be used to train a classifier with low error on the distribution, using any passive learning algorithm that receives the labeled sample as input. Therefore, from now on we focus on the problem of determining the labels of the examples in $X$.

The learner has access to a teacher, represented by an oracle $L : [m] \to \{-1, 1\}$. The goal of the learner is to find the values $L(1), \ldots, L(m)$ using as few calls to $L$ as possible.

We study active learning of the hypothesis class of halfspaces. Let $\mathbb{B}_1^d$ be the unit ball in $\mathbb{R}^d$. We assume that $X \subseteq \mathbb{B}_1^d$, and that there exists some $w^* \in \mathbb{B}_1^d$ such that $L(i) = \text{sgn}(\langle w^*, x_i \rangle)$ for all $i \in [m]$. The label complexity of an active learning algorithm is the maximal number of calls to $L$ that it makes before determining the labels of all instances in $X$, where the maximum is over all the possible functions $L$ determined by a $w^* \in \mathbb{B}_1^d$. Formally, an active learning algorithm $A$ can call $L$ several times, and then should output $(L(x_1), \ldots, L(x_m))$. Let $N(A, w^*)$ be the number of calls to $L$ that $A$ makes before outputting $(L(x_1), \ldots, L(x_m))$ if $L$ is determined by $w^*$. The (worst-case) cost of $A$ is defined as $c_{wc}(A) = \max_{w^* \in \mathbb{B}_1^d} N(A, w^*)$. We define the worst-case cost of the optimal algorithm by

$$\text{OPT} = \min_A c_{wc}(A), \tag{1}$$

where the minimum is over all active learning algorithms.

As mentioned in the introduction, our label complexity guarantees will be relative to the optimal label complexity that can be achieved for the given sample $X$. We refer to OPT as a measure of the difficulty of the active learning problem and expect our algorithm to succeed if OPT is "small". When analyzing the label complexity of our algorithm we make two relaxations. First, we further assume that $(X, L)$ is separated with a *margin* $\gamma$, that is that there exists a $w^*$ such that $\min_{i \in [m]} L(i) \langle w^*, x_i \rangle \geq \gamma$. The label complexity guarantees of our algorithm depend on $\gamma$. We describe a relaxation of this assumption of separability with a margin in Section 3.

Second, we allow our algorithm to make randomized decisions, and require it to output $(y_1, \ldots, y_m)$ such that with high probability $y_i = L(i)$ for all $i$. That is, the algorithm is allowed to fail with small probability. We will show that the number of calls to $L$ our algorithm makes is not much larger than OPT.[4]

## 3. Main results

It is easy to see that the problem of finding a policy that implements OPT for a general hypothesis class is at least as hard as the problem of finding a decision tree of minimal height. Unfortunately, this problem is NP-hard in the general case (Arkin et al., 1993). Using the additional assumption of separability with a margin, we provide an efficient algorithm that finds the correct labeling of the sample with high probability, using a number of queries which is comparable to OPT. Specifically, we prove the following theorem:

**Theorem 2** *Let $X = \{x_1, \ldots, x_m\} \subseteq \mathbb{B}_1^d$. Assume that $(X, L)$ is separable with a margin $\gamma$. Let $\delta \in (0, 1)$ be a confidence parameter. There exists a pool-based active learning algorithm with the following guarantees:*

1. *With probability at least $1 - \delta$, it returns $L(1), \ldots, L(m)$.*

2. *It requests at most $\text{OPT} \cdot 4(2d \ln(2/\gamma) + \ln(2))$ labels, where $\text{OPT}$ is defined in Equation (1).*

3. *Its running time is polynomial in $m$, $d$ and $\ln(1/\delta)$.*

The guarantees of the theorem above depend linearly on $d$, the dimension of the representation of the examples. This may be an issue in very high dimensions, and is prohibitive in the case of a kernel representation of $X$. In addition, this theorem holds only for fully separable data. These limitations can be circumvented by applying a fairly simple transformation on the points in $X$ as a preprocessing step. This transformation maps the points of $X$ to a set of points in a lower dimension, such that the new set is separable with respect to the same oracle $L$. The input points $X$ can be represented either directly as points in $\mathbb{R}^d$, or via a kernel matrix $k(x, x')$ for $x, x' \in X$. The dimension of the new representation depends on the hinge-loss of $\{(x_1, L(1)), \ldots, (x_m, L(m))\}$ with respect to the margin. Our algorithm can then be applied to the low-dimension points to retrieve the labels of $X$, with time complexity and label complexity approximation guarantees that do not depend on the original dimension $d$. As a final step, the labeled sample $\{(x_1, L(1)), \ldots, (x_m, L(m))\}$ can be used as input to a passive learner. Note that in this scheme the generalization bounds depend only on the properties of $X$ and not on the properties of the low-dimensional mapping. The following theorem formalizes the properties of the transformation.

---

4. Note that the requirements of our algorithm are easier than those of the optimal algorithm via the definition of OPT, since in the definition of $c_{wc}$ (and therefore of OPT) we do not maximize only over those $w^*$ that achieve margin $\gamma$, and we do not allow $A$ to fail with small probability. It is easy to derive cases in which OPT is large, but under the margin assumption there exists an algorithm that makes few calls to $L$. Nevertheless, OPT is a reasonable measure of the "difficulty" of the active learning problem. We leave further research on relaxed definitions of OPT to future work.

**Theorem 3** *Let $X = \{x_1, \ldots, x_m\} \subseteq B$, where $B$ is the unit ball in some Hilbert space. Let $H \geq 0$ and $\gamma > 0$, and assume there exists a $w^* \in B$ such that*

$$H \geq \sum_{i=1}^{m} \max(0, \gamma - L(i)\langle w^*, x_i \rangle)^2.$$

*Let $\delta \in (0, 1)$ be a confidence parameter. There exists an algorithm that receives $X$ as vectors in $\mathbb{R}^d$ or as a kernel matrix $K \in \mathbb{R}^{m \times m}$, and input parameters $\gamma$ and $H$, and outputs a set $\bar{X} = \{\bar{x}_1, \ldots, \bar{x}_m\} \subseteq \mathbb{R}^k$, such that*

1. *$k = O\left(\frac{(H+1)\ln(m/\delta)}{\gamma^2}\right)$,*

2. *With probability $1 - \delta$, $\bar{X} \subseteq \mathbb{B}_1^k$ and $(\bar{X}, L)$ is separable with a margin $\frac{\gamma}{2+2\sqrt{H}}$.*

3. *The run-time of the algorithm is polynomial in $d, m, 1/\gamma, \ln(1/\delta)$ if $x_i$ are represented as vectors in $d$, and is polynomial in $m, 1/\gamma, \ln(1/\delta)$ if $x_i$ are represented by a kernel matrix.*

Finally, consider the common learning setting in which there is some distribution $D$ over (instance,label) pairs, and $\{(x_1, L(1)), \ldots, (x_m, L(m))\}$ are i.i.d. pairs drawn from $D$. If $D$ is separable with a margin $\gamma$, then it suffices to have a labeled sample of size $m = O(\frac{1}{\gamma^2 \epsilon})$ to allow passive learning to accuracy $\epsilon$. The transformation above thus maps the sample to dimension $O(\frac{1}{\gamma^2} \ln(\frac{1}{\gamma^2 \epsilon \delta}))$. Executing our algorithm on the resulting sample we get an active learning algorithm that runs in time polynomial in $\frac{1}{\gamma}$, $\frac{1}{\epsilon}$ and $\ln(1/\delta)$, and has a label complexity which is guaranteed to be no more than $\text{OPT} \cdot O(\frac{1}{\gamma^2} \ln(\frac{1}{\gamma^2 \epsilon \delta}))$, where OPT is the number of queries required by an optimal active learning algorithm on the transformed sample.

## 4. The ALuMA algorithm

In this section we describe our algorithm. We name it `Active Learning under a Margin Assumption` or ALuMA for short.

To explain our approach, it is convenient to think about the active learning process as a search procedure for the vector $w^*$ that determines the labels $L(1), \ldots, L(m)$. Suppose our first $t$ calls to $L$ are for the labels $L(i_1), \ldots, L(i_t)$, and denote $P_t = \{(x_{i_1}, L(i_1)), \ldots, (x_{i_t}, L(i_t))\}$. Then, we know that $w^*$ must be in the set

$$V(P_t) = \{w \in \mathbb{B}_1^d : \forall (x, y) \in P_t, \ y\langle w, x \rangle > 0\} \ .$$

The set $V(P_t)$ is called the *version space* induced by $P_t$.

Intuitively, we would like to query labels that will make $V(P_t)$ "small". There are many ways to define the "size" of $V(P_t)$. One way, which is convenient for our analysis, is to define the size of a version space by its volume, denoted $\mathrm{Vol}(V(P_t))$. Therefore, ideally, we would like to query the labels $L(i_1), \ldots, L(i_t)$ for which $\mathrm{Vol}(V(P_t))$ is minimal. Naturally, the size of $V(P_t)$ depends on the actual labels we will receive from $L$, which are unknown to us prior to querying them. We therefore would like to query the labels for which $\mathrm{Vol}(V(P_t))$ is minimal in the worst-case, where the worst-case is over all possible vectors $w^*$ that may determine $L(1), \ldots, L(m)$.

The number of all possible sequences of $t$ queries grows exponentially with $t$, which poses a computational challenge. We thus follow a simple greedy approach: At each iteration, query the example whose label will lead to a version space of minimal size. Again, since we do not know the label prior to querying, we would like to choose the example which leads to the minimal size in the worst-case, where the worst-case is over all possible vectors $w$ in the current version space. Formally, given the current version space $V(P_t)$, the next query should be for the label of the example in

$$\operatorname*{argmin}_{x \in X} \ \max_{w \in V(P_t)} \ \mathrm{Vol}(V(P_t \cup \{(x, \mathrm{sgn}(\langle w, x \rangle))\})) \ . \tag{2}$$

Denoting $V_{t,x}^1 = V(P_t \cup \{(x, 1)\})$ and $V_{t,x}^{-1} = V(P_t \cup \{(x, -1)\})$, an equivalent way to write Equation (2) is [5]

$$\operatorname*{argmax}_{x \in X} \ \mathrm{Vol}(V_{t,x}^1) \cdot \mathrm{Vol}(V_{t,x}^{-1}) \ . \tag{3}$$

To implement Equation (3), we need to be able to calculate the volumes of the sets $V_{t,x}^1$ and $V_{t,x}^{-1}$. Both of these sets are convex sets obtained by intersecting the unit ball with halfspaces. The problem of calculating the volume of such convex sets in $\mathbb{R}^d$ is #P-hard if $d$ is not fixed (Brightwell and Winkler, 1991). Moreover, deterministically approximating the volume is NP-hard in the general case (Matoušek, 2002). Luckily, it is possible to approximate this volume using randomization. Specifically, in Kannan et al. (1997) a randomized algorithm with the following guarantees is provided:

**Lemma 4** *Let $K \subseteq \mathbb{R}^d$ be a convex body with an efficient separation oracle. There exists a randomized algorithm, such that given $\epsilon, \delta > 0$, with probability at least $1 - \delta$ the algorithm*

---

5. To see the equivalence, denote $\alpha(x) = \max_{w \in V(P_t)} \mathrm{Vol}(V_{t,x}^{\mathrm{sgn}(\langle w, x \rangle)})/\mathrm{Vol}(V(P_t))$. Clearly, Equation (2) can be written as $\operatorname{argmin}_x \alpha(x)$. Note that $\alpha(x) \in [1/2, 1]$ and $1 - \alpha(x) = \mathrm{Vol}(V_{t,x}^{-\mathrm{sgn}(\langle w, x \rangle)})/\mathrm{Vol}(V(P_t))$. Therefore, Equation (3) can be written as

$$\operatorname*{argmax}_x \alpha(x)(1 - \alpha(x)) = \operatorname*{argmax}_x \min\{\alpha(x), 1 - \alpha(x)\} = \operatorname*{argmin}_x \max\{1 - \alpha(x), \alpha(x)\} \ .$$

Since $\alpha(x) \in [1/2, 1]$ we conclude that the above equals $\operatorname{argmin}_x \alpha(x)$ as required.

*returns a non-negative number $\Gamma$ such that $(1 - \epsilon)\Gamma < \mathbb{P}(K) < (1 + \epsilon)\Gamma$. The running time of the algorithm is polynomial in $d, 1/\epsilon, \ln(1/\delta)$.*

ALuMA uses this algorithm to estimate $\text{Vol}(V_{t,x}^1)$ and $\text{Vol}(V_{t,x}^{-1})$ with sufficient accuracy. We denote an execution of this algorithm on a convex body $K$ by $\Gamma \leftarrow \text{VolEst}(K, \epsilon, \delta)$. The convex body $K$ is represented in the algorithm by the set of the constraints that define it.

ALuMA terminates when it exhausts its budget of queries, which is provided as a parameter to the algorithm. If the final version space $V$ does not determine the labeling of $X$, ALuMA randomly draws several hypotheses from $V$ and labels $X$ according to a majority vote over these hypotheses. Our analysis will show that if the budget is large enough and the draw of the hypotheses is approximately uniform from $V$, then this strategy leads to the correct labeling of $X$ with high probability.

To draw a hypothesis approximately uniformly from $V$, we use the hit-and-run algorithm (Lovász, 1999), which draws a random sample from a convex body $K$ according to a distribution which is close in total variation distance to the uniform distribution over $K$. Formally, The following definition parametrizes the closeness of a distribution to the uniform distribution:

**Definition 5** *Let $K \subseteq \mathbb{R}^d$ be a convex body with an efficient separation oracle, and let $\tau$ be a distribution over $K$. $\tau$ is $\lambda$-uniform if $\sup_A |\tau(A) - \mathbb{P}(A)/\mathbb{P}(K)| \leq \lambda$, where the supremum is over all measurable subsets of $K$.*

The hit-and-run algorithm draws a sample from a $\lambda$-uniform distribution in time $\tilde{O}(d^3/\lambda^2)$.

ALuMA is listed below as Alg. 1. Its inputs are the unlabeled sample $X$, the labeling oracle $L$, the maximal allowed number of label queries $N$, and the desired confidence $\delta \in (0, 1)$. It returns the labels of all the examples in $X$.

---

**Algorithm 1** The **ALuMA** algorithm

---
1: **Input:** $X = \{x_1, \ldots, x_m\}$, $L : [m] \rightarrow \{-1, 1\}$, $N$, $\delta$
2: $I_1 \leftarrow [m]$
3: $V_1 \leftarrow \mathbb{B}_1^d$
4: **for** $t = 1$ to $N$ **do**
5:     $\forall i \in I_t, j \in \{\pm 1\}$, do $\hat{v}_{x_i, j} \leftarrow \text{VolEst}(V_{t,x_i}^j, \frac{1}{3}, \frac{\delta}{4mN})$
6:     Select $i_t \in \text{argmax}_{i \in I_t}(\hat{v}_{x_i, 1} \cdot \hat{v}_{x_i, -1})$
7:     $I_{t+1} \leftarrow I_t \setminus \{i_t\}$
8:     Request $y = L(i_t)$
9:     $V_{t+1} \leftarrow V_t \cap \{w : y\langle w, x_{i_t}\rangle > 0\}$
10: **end for**
11: $M \leftarrow \lceil 72 \ln(2/\delta) \rceil$.
12: Draw $w_1, \ldots, w_M$ $\frac{1}{12}$-uniformly from $V_{N+1}$.
13: For each $x_i$ return the label $y_i = \text{sgn}\left(\sum_{j=1}^{M} \text{sgn}(\langle w_j, x_i \rangle)\right)$.

---

## 5. Proof outline

In this section we describe the outline of the proof of Theorem 2. The detailed proof is given in Appendix A. Given $S \subseteq X$ and $w$, we define the partial realization of $w$ on $S$ as

$$w|_S = \{(x, \mathrm{sgn}(\langle w, x \rangle)) : x \in S\} \ .$$

Any active learning algorithm works as follows. Let $w$ be a vector that determines $L$, which is unknown to the learner. The algorithm starts with $S_1 = \emptyset$. At iteration $t$, the algorithm knows $w|_{S_t}$, and based on this information, it selects a new example $x \in X$ and sets $S_{t+1} = S_t \cup \{x\}$. We can therefore represent any algorithm by a policy function $\pi$ which maps each partial realization to an element from $X$.[6] We denote by $S(\pi, w, k)$ the value of $w|_{S_k}$ if policy $\pi$ is applied for $k$ iterations, and the received labels are consistent with $w$.

We define a utility function over partial realizations as follows. Let $U$ be the uniform distribution over $\mathbb{B}_1^d$. That is, a measurable subset $Z \subseteq \mathbb{B}_1^d$ has probability mass $U(Z) = \mathrm{Vol}(Z)/\mathrm{Vol}(\mathbb{B}_1^d)$. Given a partial realization $w|_S$ for some $S \subseteq X$ and $w \in \mathbb{B}_1^d$, we define the utility of the partial realization to be

$$f(w|_S) = 1 - U(V(w|_S)) = \mathop{\mathbb{P}}_{v \sim U}[v \notin V(w|_S)] \ . \tag{4}$$

That is, $f$ measures the probability mass of all vectors in $\mathbb{B}_1^d$ which are not in the version space corresponds to the partial realization. Intuitively, a good policy should yield partial realizations of high utility.

Given a budget of $k$ calls to the labeling oracle, we would like to construct a policy $\pi$ for which $f(S(\pi, w^*, k))$ is as large as possible, in the worst-case over the choice of $w^*$. For technical reasons, we prefer to derive a policy aiming to maximize the expected value of $f(S(\pi, w^*, k))$, where expectation is with respect to $w^* \sim U$. That is, we define

$$f_{\mathrm{avg}}(\pi, k) = \mathbb{E}_{w^* \sim U}[f(S(\pi, w^*, k))] \ .$$

With this definition at hand, a non-efficient approach for deriving a good policy function is to perform exhaustive search over all policies, and then choose the one for which $f_{\mathrm{avg}}(\pi, k)$ is maximal. On the other hand we show that ALuMA, which is an efficient algorithm, implements an approximated greedy policy for increasing $f_{\mathrm{avg}}(\pi, k)$. Indeed, let $P_t$ be the partial realization achieved by ALuMA by the beginning of iteration $t$. At this point the algorithm "knows" that the correct separator is in $V(P_t)$. Therefore, to make $f_{\mathrm{avg}}(\texttt{ALuMA}, k)$ larger, we perform an approximately greedy step, by querying the label of an example which approximately maximizes

$$\mathop{\mathbb{E}}_{w^* \sim U}[f(P_t \cup \{(x, \mathrm{sgn}(\langle w^*, x \rangle))\}) \mid w^* \in V(P_t)] \ .$$

Standard algebraic manipulations can show that the above is equivalent to Equation (3).

We see that ALuMA performs an approximated greedy policy for maximizing $f_{\mathrm{avg}}$. But, how good is this greedy policy? Recently, Golovin and Krause (2010) showed that if $f$ satisfies certain conditions then an approximately greedy policy achieves an approximately-optimal expected utility. These conditions are called *adaptive submodularity* and *adaptive*

---

6. For randomized algorithms, we can define a policy function for any realization of their random bits.

*monotonicity*, and it can be shown that our utility function adheres to these conditions. We use this to show (see Section A.1) that for any policy $\pi$ and integer $k$, if we run ALuMA for $n$ iterations then

$$f_{\text{avg}}(\text{ALuMA}, n) \geq f_{\text{avg}}(\pi, k) - e^{-\frac{n}{4k}} \ . \tag{5}$$

In particular, this holds for the policy $\pi^*$ that implements the optimal algorithm in the definition of OPT, and for $k = \text{OPT}$. Since for any $w$, $S(\pi^*, w, \text{OPT})$ determines the predictions of $w$ on all the examples in $X$, it follows that $V(S(\pi^*, w, \text{OPT})) \subseteq V(S(\text{ALuMA}, w, n))$. This fact can be used to show (see Lemma 12) that for any $w$,

$$f_{\text{avg}}(\pi^*, \text{OPT}) - f_{\text{avg}}(\text{ALuMA}, n) \geq U(V(w|_X)) \left( U(V(S(\text{ALuMA}, w, n))) - U(V(w|_X)) \right) \ .$$

Combining the above with Equation (5) and rearranging terms yields

$$\forall w, \quad \frac{U(V(w|_X))}{U(V(S(\text{ALuMA}, w, n)))} \geq \frac{U(V(w|_X))^2}{e^{-\frac{n}{4k}} + U(V(w|_X))^2}. \tag{6}$$

Our final step is to use the assumption that that $w$ separates $X$ with margin $\gamma$, which implies that $U(V(w|_X)) \geq (\gamma/2)^d$ (Lemma 13). Plugging this into Equation (6) yields that, for $n$ large enough, at least $2/3$ of the vectors in $V(S(\text{ALuMA}, w, n))$ are also in $V(w|_X)$. Hence, an (approximate) majority vote over $V(S(\text{ALuMA}, w, n))$ will correctly determine the labels of all the examples in $X$ (Corollary 15).

## 6. Handling inseparable data, high-dimensions, or kernels

If the data $X = \{x_1, \ldots, x_m\}$ is in a very high dimension, or it is not guaranteed to be separable, or it is represented only using a kernel matrix, then ALuMA still can be used, after a preprocessing step. This preprocessing step maps the points in $X$ to a set of points in a lower dimension, which are separable using the original labels of $X$. We describe the procedure below, and prove that it satisfies the requirements of Theorem 3 in Appendix B.

The preprocessing step is composed of two simple transformations. In the first transformation, which can be skipped if the data is known to be separable, each example $x_i \in X$ is mapped to an example in dimension $d + m$, defined by $x_i' = (ax_i; \sqrt{1 - a^2} \cdot e_i)$, where $e_i$ is the $i$'th vector of the natural basis of $\mathbb{R}^m$ and $a > 0$ is a scalar that will be defined below. Thus the first $d$ coordinates of $x_i'$ hold the original vector times $a$, the rest of the coordinates are zero,except for $x_i'[d + i] = \sqrt{1 - a^2}$. This mapping guarantees that the set $X' = (x_1', \ldots, x_m')$ is separable with the same labels as those of $X$, and with a margin that depends on the cumulative squared-hinge-loss of the data.

In the second transformation, a Johnson-Lindenstrauss random projection (Johnson and Lindenstrauss, 1984; Bourgain, 1985) is applied to $X'$, thus producing a new set of points $\bar{X} = (\bar{x}_1, \ldots, \bar{x}_m)$ in a lower dimension $\mathbb{R}^k$, where $k$ depends on the original margin and on the amount of margin error. With high probability, the new set of points will be separable with a margin that also depends on the original margin and on the amount of margin error.

If the input data is provided not as vectors in $\mathbb{R}^d$ but via a kernel matrix, then a simple decomposition is performed before the preprocessing begins. The full preprocessing procedure is listed below as Alg. 2. The first input to the algorithm is the data for preprocessing, given as $X \subseteq \mathbb{R}^d$ or as a kernel matrix $K \in \mathbb{R}^{m \times m}$. The other inputs are $\gamma$—a margin parameter, $H$—an upper bound on the margin error relative to $\gamma$, and $\delta$, which is the required confidence.

---

**Algorithm 2** Preprocessing

---

1: **Input:** $X = \{x_1, \ldots, x_m\}$ or $K \in \mathbb{R}^{m \times m}$, $\gamma$, $H$, $\delta$
2: **if** input data is a kernel matrix $K$ **then**
3:      Find $U \in \mathbb{R}^{m \times m}$ such that $K = UU^T$
4:      $\forall i \in [m], x_i \leftarrow$ row $i$ of $U$
5:      $d \leftarrow m$
6: **end if**
7: $a \leftarrow \sqrt{\frac{1}{1 + \sqrt{H}}}$
8: $\forall i \in [m], x_i' \leftarrow (ax_i; \sqrt{1 - a^2} \cdot e_i)$
9: $k \leftarrow O\left(\frac{(H+1)\ln(m/\delta)}{\gamma^2}\right)$
10: $M \leftarrow$ a random $\{\pm 1\}$ matrix of dimension $k \times (d + m)$
11: **for** $i \in [m]$ **do**
12:      $\bar{x}_i \leftarrow Mx_i'$
13: **end for**
14: Return $(\bar{x}_1, \ldots, \bar{x}_m)$.

---

After the preprocessing step, $\bar{X}$ is used as input to ALuMA, which then returns a set of labels for the examples in $\bar{X}$. These are also the labels of the examples in the original $X$. To retrieve a halfspace for $X$ with the least margin error, any passive learning algorithm can be applied to the resulting labeled sample. The full active learning procedure is described in Alg. 3.

Note that if ALuMA returns the correct labels for the sample, the usual generalization bounds for passive supervised learning can be used to bound the true error of the returned separator $w$. In particular, we can apply the support vector machine algorithm (SVM) and rely on generalization bounds for SVM.

---

**Algorithm 3** Active Learning

---

1: **Input:** $X = \{x_1, \ldots, x_m\}$ or $K \in \mathbb{R}^{m \times m}$, $L : [m] \rightarrow \{-1, 1\}$, $N$, $\gamma$, $H$, $\delta$
2: **if** input has $X$ **then**
3:      Get $\bar{X}$ by running Alg. 2 with input $X, \gamma, H, \delta/2$.
4: **else**
5:      Get $\bar{X}$ by running Alg. 2 with input $K, \gamma, H, \delta/2$.
6: **end if**
7: Get $(y_1, \ldots, y_m)$ by running ALuMA with input $\bar{X}, L, N, \delta/2$.
8: Get $w \in \mathbb{R}^d$ by running SVM on the labeled sample $\{(x_1, y_1), \ldots, (x_m, y_m)\}$.
9: Return $w$.

---

## 7. A Simpler Implementation of ALuMA

The ALuMA algorithm described in Alg. 1 uses $O(Nm)$ volume estimations as a black-box procedure, where $N$ is the budget of labels and $m$ is the pool size. The complexity of each application of the volume estimation procedure is $\tilde{O}(d^5)$ where $d$ is the dimension. Thus the overall complexity of the algorithm is $\tilde{O}(Nmd^5)$. This complexity can be somewhat improved under some "luckiness" conditions.

The volume estimation procedure uses $\lambda$-uniform sampling based on hit-and-run as its core procedure. Instead, we can use hit-and-run directly as follows: At each iteration of ALuMA, instead of step 5, perform the following procedure:

---
**Algorithm 4** Estimation Procedure
---
1: Input: $\lambda \in (0, \frac{1}{24}), V_t, I_t$
2: $k \leftarrow \frac{\ln(2Nm/\delta)}{2\lambda^2}$
3: Sample $h_1, \ldots, h_k \in V_t$ $\lambda$-uniformly.
4: $\forall i \in I_t, j \in \{-1, +1\}, \hat{v}_{x_i,j} \leftarrow \frac{1}{k} |\{i \mid h_i(x_i) = j\}|$.
---

The complexity of ALuMA when using this procedure is $\tilde{O}(N(d^3/\lambda^4 + m/\lambda^2))$, which is better than the complexity of the full Alg. 1 for a constant $\lambda$. An additional practical benefit of this alternative estimation procedure is that when implementing, it is easy to limit the actual computation time used in the implementation by running the procedure with a smaller number $k$ and a smaller number of hit-and-run mixing iterations.[7] This provides a natural trade-off between computation time and labeling costs.

Theorem 20 in Appendix C shows that letting ALuMA use Alg. 4 as the estimation procedure results in similar guarantees to those of the original implementation of ALuMA (Alg. 1). The only condition is that the best example in each iteration induces a fairly balanced partition of the current version space. In our experiments we noticed that this is generally the case in practice. Moreover, the theorem shows that it is possible to verify that the condition holds while running the algorithm. Thus, the estimation procedure can easily be augmented with an additional verification step at the beginning of each iteration. On iterations that fail the verification, the algorithm will use the original black-box volume estimation procedure.

## 8. Experiments

We evaluated ALuMA over synthetic and real data sets and compared its label complexity performance to that of a passive ERM (that is, one that uses random labeled examples), as well as to that of QBC and CAL.

Our implementation of ALuMA uses hit-and-run samples instead of full-blown volume estimation, as detailed in Section 7 above. QBC is also implemented using hit-and-run as in Gilad-Bachrach et al. (2005). For both ALuMA and QBC, we used a fixed number of mixing iterations for hit-and-run, which we set to 1000. We also fixed the number of sampled hypotheses at each iteration of ALuMA to 1000, and used the same set of hypotheses to calculate the majority vote for classification. CAL and QBC examine the examples sequentially, thus the input provided to them was a random ordering of the example pool. Since the active learners operate by reducing the training error, the graphs we show compare the training errors of the different algorithms. The test errors show a similar trend.

---

7. Gilad-Bachrach et al. (2005) report that the actual mixing time of hit-and-run is much faster than the one guaranteed by the theoretical bounds, and we have observed a similar phenomenon in our experiments.

| $d$ | ALuMA | QBC | CAL | ERM |
|---|---|---|---|---|
| 10 | 29 | 50 | 308 | 1008 |
| 12 | 38 | 113 | 862 | 3958 |
| 15 | 55 | 150 | 2401 | $> 20000$ |

Table 1: Octahedron: Number of iterations to achieve zero error

In each of the algorithms, the classification of the training examples is done using the version space defined by the queried labels. The theory for CAL and ERM allows selecting an arbitrary predictor out of the version space. In QBC, the hypothesis should be drawn uniformly at random from the version space. However, we have found that all the algorithms show a significant improvement in classification error if they classify using the majority vote classification proposed for ALuMA. Therefore, in all of our experiments, the results for all the algorithms are based on a majority vote classification.

The first experiment is synthetic: the pool of examples is taken to be the support of the distribution described in Example 2 (the octahedron example), with an additional dimension to account for halfspaces with a bias. We also added the negative vertices $-e_i$ to the pool. Similarly to the proof of Theorem 1, it suffices to query the vertices of the octahedron to reach zero error. Table 1 lists the number of iterations required in practice to achieve zero error by each of the algorithms. ALuMA is clearly much better than QBC and CAL. Furthermore, the number of queries ALuMA requires is indeed close to the number of vertices.

The second batch of experiments is with the MNIST dataset.[8] The examples in this dataset are gray-scale images of handwritten digits in dimension 784. Each digit has about $6,000$ training examples. We performed binary active learning by pre-selecting pairs of digits.

Figure 1 and Figure 2 depict the training error as a function of the label budget when learning to distinguish the digits 3 and 5, and between the digits 4 and 7. Both these digits pairs are linearly separable in this dataset. Figure 1 depicts the error as a function of the label budget. It is striking to observe that CAL provides no improvement over passive ERM in the first 1000 examples, while this budget suffices to reach zero training error for ALuMA.

We also tested the effectiveness of our approach for data with labeling errors (see Section 6). To this end we applied the preprocessing algorithm listed in Alg. 2 to the linear representation of the digits 2 and 3, which are not separable in the original representation. Applying the Johnson-Lindenstrauss projection to the enhanced representation resulted in a separable representation in dimension 800. Figure 3 depicts the performance of ALuMA, CAL, QBC and ERM, all on the transformed separable representation. Finally, to test the use of kernel representations, we generated a dataset in which two digits (4 and 7) are labeled as positive, and two other digits (3 and 5) are labeled as negative, and used a kernel

---

8. http://yann.lecun.com/exdb/mnist/

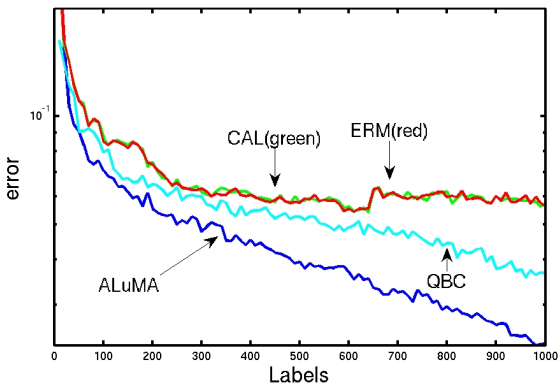

Figure 1: MNIST 3 vs. 5

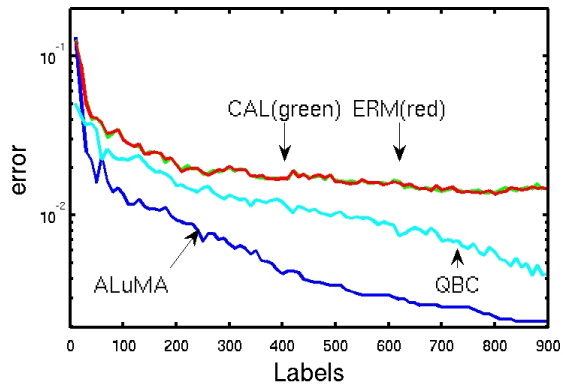

Figure 2: MNIST 4 vs. 7

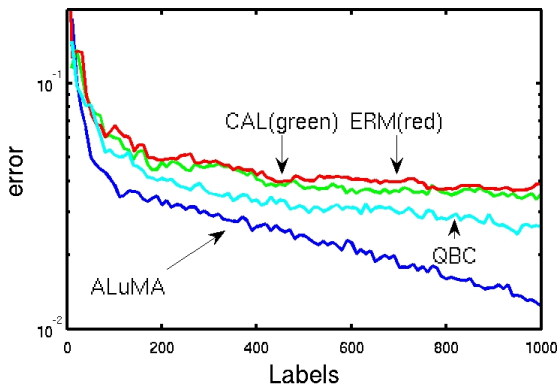

Figure 3: MNIST 2 vs. 3 (non-separable).

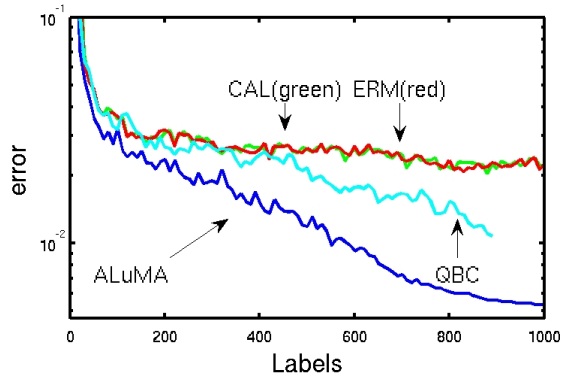

Figure 4: MNIST 4,7 vs. 3,5 with kernel RBF.

RBF representation so that the data is separable. The preprocessing step resulted in a separable representation in dimension 700. The results of running each of the algorithms on this representation are depicted in Figure 4. Note that QBC does not use the entire budget of labels in this experiment. This is because the mechanism by which QBC selects an example takes a very long time when the training error is small, thus we were not able to run it long enough so that it uses its full label budget.

We also tested the algorithms on the PCMAC dataset.[9] This is a real-world data set, which represents a two-class categorization of the 20-Newsgroup collection. The examples are web-posts represented using bag-of-words. The original dimension of examples is 7511. We used the Johnson-Lindenstrauss projection to reduce the dimension to 300, which kept the data still separable. We used a training set of 1000 examples. Figure 5 depicts the results. Here too we were not able to run QBC long enough to use its entire budget.

The experiments on MNIST and PCMAC show that ALuMA is superior to CAL and QBC for real-world distributions, in which CAL and QBC have no theoretical analysis. The next experiment shows that ALuMA outperforms CAL and QBC even on a data sampled

9. http://vikas.sindhwani.org/datasets/lskm/matlab/pcmac.mat

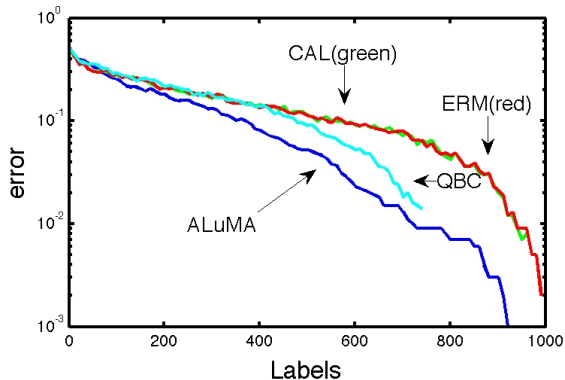

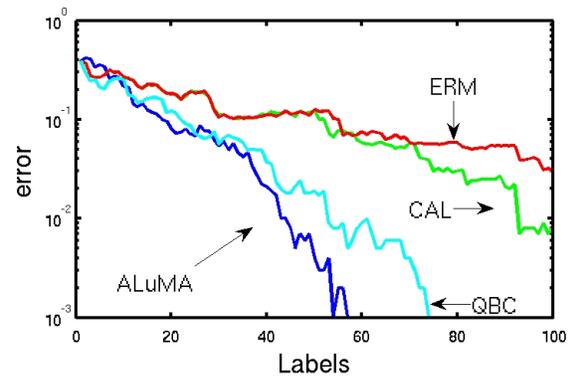

Figure 5: PCMAC

Figure 6: Uniform distribution ($d = 10$).

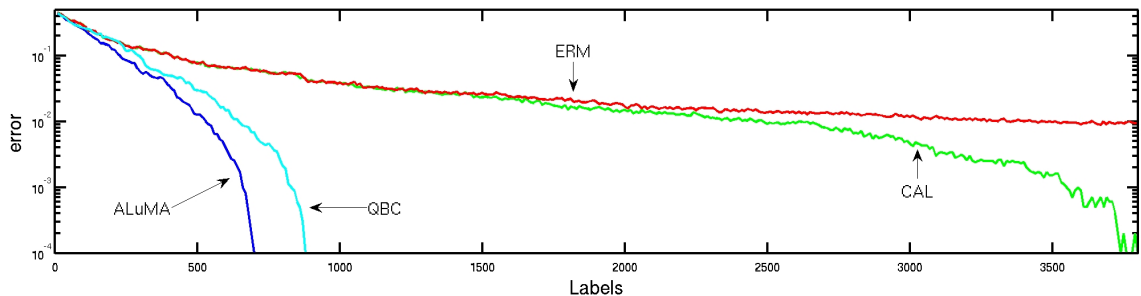

Figure 7: Uniform distribution ($d = 100$).

from the uniform distribution on a sphere, which is the distribution with the best guarantees for both CAL and QBC. Figure 6 and Figure 7 depict the training error as a function of the label budget when learning a random halfspace over the uniform distribution in $\mathbb{R}^{10}$ and in $\mathbb{R}^{100}$. Here too, CAL requires many more labels than ALuMA requires. For $\mathbb{R}^{100}$, it improves over passive learning only long after ALuMA has reached zero training error. The difference between the performance of the different algorithms is less marked for $d = 10$ than for $d = 100$, suggesting that the difference grows with the dimension.

## 9. Discussion

We studied active learning of halfspaces under a margin assumption. We have shown that a large margin assumption alone cannot guarantee a uniform improvement in label complexity over passive learning. However, the margin assumption enables us to derive an algorithm with regret-type guarantees: it will not query many more labels than OPT, where OPT is the minimal number of queries required to achieve zero training error on the given pool of examples, in the worst-case over the target hypotheses.

An open problem, which we leave for future work, is whether one can be efficiently competitive with respect to a relaxed definition of OPT, one which only requires achieving zero training error with high probability, and only if the target hypothesis separates the training set with a margin $\gamma$.

# References

E.M. Arkin, H. Meijer, J.S.B. Mitchell, D. Rappaport, and S.S. Skiena. Decision trees for geometric models. In *Proceedings of the ninth annual symposium on Computational geometry*, pages 369–378. ACM, 1993.

M.F. Balcan, A. Beygelzimer, and J. Langford. Agnostic active learning. In *Proceedings of the 23rd international conference on Machine learning*, pages 65–72. ACM, 2006a.

M.F. Balcan, A. Blum, and S. Vempala. Kernels as features: On kernels, margins, and low-dimensional mappings. *Machine Learning*, 65(1):79–94, 2006b.

M.F. Balcan, A. Broder, and T. Zhang. Margin based active learning. *Learning Theory*, pages 35–50, 2007.

A. Beygelzimer, S. Dasgupta, and J. Langford. Importance weighted active learning. In *Proceedings of the 26th Annual International Conference on Machine Learning*, pages 49–56. ACM, 2009.

J. Bourgain. On lipschitz embedding of finite metric spaces in hilbert space. *Israel Journal of Mathematics*, 52(1):46–52, 1985.

G. Brightwell and P. Winkler. Counting linear extensions is #p-complete. In *Proceedings of the twenty-third annual ACM symposium on Theory of computing*, STOC '91, pages 175–181, 1991.

D. Cohn, L. Atlas, and R. Ladner. Improving generalization with active learning. *Machine Learning*, 15(2):201–221, 1994.

N. Cristianini and J. Shawe-Taylor. *Kernel Methods for Pattern Analysis*. Cambridge University Press, 2004.

S. Dasgupta. Coarse sample complexity bounds for active learning. *Advances in neural information processing systems*, 18:235, 2006.

S. Dasgupta, A. Kalai, and C. Monteleoni. Analysis of perceptron-based active learning. *Learning Theory*, pages 889–905, 2005.

Y. Freund, H.S. Seung, E. Shamir, and N. Tishby. Selective sampling using the query by committee algorithm. *Machine learning*, 28(2):133–168, 1997.

E. Friedman. Active learning for smooth problems. In *Proceedings of the 22nd Conference on Learning Theory*, volume 1, pages 3–2, 2009.

R. Gilad-Bachrach, A. Navot, and N. Tishby. Query by committee made real. *Advances in Neural Information Processing Systems (NIPS)*, 19, 2005.

D. Golovin and A. Krause. Adaptive submodularity: A new approach to active learning and stochastic optimization. In *Proceedings of International Conference on Learning Theory (COLT)*, 2010.

S. Hanneke. Teaching dimension and the complexity of active learning. In *COLT*, 2007.

S. Hanneke. Adaptive rates of convergence in active learning. In *COLT*, 2009.

W. Johnson and J. Lindenstrauss. Extensions of lipschitz mapping into hilbert space. *Contemporary Mathematics*, 26:189–206, 1984.

R. Kannan, L. Lovász, and M. Simonovits. Random walks and an $o*(n^5)$ volume algorithm for convex bodies. *Random structures and algorithms*, 11(1):1–50, 1997.

L. Lovász. Hit-and-run mixes fast. *Mathematical Programming*, 86(3):443–461, 1999.

J. Matoušek. *Lectures on discrete geometry*, volume 212. Springer Verlag, 2002.

A. McCallum and K. Nigam. Employing em in pool-based active learning for text classification. In *Proceedings of ICML-98, 15th International Conference on Machine Learning*, pages 350–358, 1998.

Sivan Sabato, Nathan Srebro, and Naftali Tishby. Tight sample complexity of large-margin learning. In *Advances in Neural Information Processing Systems 23 (NIPS)*, pages 2038–2046, 2010.

B. Schölkopf and A. J. Smola. *Learning with Kernels: Support Vector Machines, Regularization, Optimization and Beyond*. MIT Press, 2002.

H.S. Seung, M. Opper, and H. Sompolinsky. Query by committee. In *Proceedings of the fifth annual workshop on Computational learning theory*, pages 287–294. ACM, 1992.

C.E. Shannon. Probability of error for optimal codes in a gaussian channel. *Bell System Technical Journal*, 38:611–656, 1959.

S. Tong and D. Koller. Support vector machine active learning with applications to text classification. *The Journal of Machine Learning Research*, 2:45–66, 2002.

# Appendix A. Analysis of ALuMA

In this section we prove Theorem 2 by showing that ALuMA satisfies the conditions of the theorem. First, note that each step in Alg. 1 runs in polynomial time in $N, m, d$ and $\ln(1/\delta)$. Since each step is repeated at most $N \leq m$ times, ALuMA is polynomial in $m, d$ and $\ln(1/\delta)$. This proves item (3) of Theorem 2. In the following we prove items (1) and (2), following the proof outline described in Section 5. The result will be stated formally as Corollary 15.

## A.1 ALuMA Increases the Utility Function Fast

Recall that in Equation (4) we defined a utility function $f$, that measures the progress of our algorithm. In this section we redefine $f$ using a slightly different notation. Let $\mathcal{H}$ be the hypothesis class induced by homogeneous halfspaces in $\mathbb{R}^d$. We define a version space of a partial realization as a subset of $\mathcal{H}$ as follows:

$$V(P_t) = \{h \in \mathcal{H} : \forall (x, y) \in P_t, \ h(x) = y\} \ .$$

We further define the probability mass of a set $G \subseteq \mathcal{H}$ by

$$\mathbb{P}(G) = \mathop{\mathbb{P}}_{w \sim U}(\{w \in \mathbb{B}_1^d \mid \exists h \in G, \forall x \in \mathbb{R}^d, \ h(x) = \mathrm{sgn}(\langle w, x \rangle)\}),$$

where $U$ is the uniform distribution over $\mathbb{B}_1^d$. Consequently, the expected value of a random variable, $g : \mathcal{H} \to \mathbb{R}$, is defined as

$$\mathbb{E}_h[g(h)] = \int_{h \in \mathcal{H}} g(h) \, \mathbb{P}(dh) \ .$$

The utility function $f$ for $h \in \mathcal{H}$ and $S \subseteq X$ is thus

$$f(h|_S) = 1 - \mathbb{P}(V(h|_S)) = \mathbb{P}(\mathcal{H} \setminus V(h|_S)). \tag{7}$$

Let $\mathcal{L}_{X,\mathcal{H}} = \{h|_{X'} : X' \subseteq X, h \in \mathcal{H}\}$ be the set of all possible partial labelings of $X$ by a hypothesis in $\mathcal{H}$. A policy is any function $\pi : \mathcal{L}_{X,\mathcal{H}} \to X$. For a policy $\pi$, an integer $k$, and a hypothesis $h \in \mathcal{H}$, we denote by $S(\pi, h, k)$ the first $k$ (example,label) pairs queried by $\pi$, under the assumption that $L = h|_X$. That is,

$$S(\pi, h, 1) = \{(\pi(\emptyset), h(\pi(\emptyset)))\}, \text{ and}$$
$$S(\pi, h, k) = S(\pi, h, k-1) \cup \{(\pi(S(\pi, h, k-1)), h(\pi(S(\pi, h, k-1))))\}.$$

The expected utility of applying policy $\pi$ for $k$ steps is

$$f_{\mathrm{avg}}(\pi, k) = \mathbb{E}_h[f(S(\pi, h, k))].$$

In this section we show that ALuMA increases $f_{\mathrm{avg}}$ almost as fast as any other policy, including the optimal one. We prove that with probability at least $1 - \delta/2$, for any policy $\pi$ and any $n, k > 0$,

$$f_{\mathrm{avg}}(\texttt{ALuMA}, n) \geq f_{\mathrm{avg}}(\pi, k) - e^{-\frac{n}{4k}}. \tag{8}$$

To prove this inequality we present the notion of an adaptive sub-modular function, first defined in Golovin and Krause (2010). Let $f : \mathcal{L}_{X,\mathcal{H}} \to \mathbb{R}_+$ be any utility function from the set of possible partial labelings of $X$ to the non-negative reals. We define the notions of *adaptive monotonicity* and *adaptive submodularity* of a utility function using the following notation: For an element $x \in X$, a subset $Z \subseteq X$ and a hypothesis $h \in \mathcal{H}$, we define the conditional expected marginal benefit of $x$, conditioned on having observed the partial labeling $h|_Z$, by

$$\Delta(h|_Z, x) = \mathbb{E}_g\big[f(g|_{Z \cup \{x\}}) - f(g|_Z) \, \big| \, g|_Z = h|_Z\big].$$

Put another way, $\Delta(h|_Z, x)$ is the expected improvement of $f$ if we add to $Z$ the element $x$, where expectation is over a choice of a hypothesis $g$ taken uniformly at random from the set of hypotheses that agree with $h$ on $Z$.

**Definition 6 (Adaptive Monotonicity)** *A utility function $f : \mathcal{L}_{X,\mathcal{H}} \to \mathbb{R}_+$ is adaptive monotone if the conditional expected marginal benefit is always non-negative. That is, if for all $h \in \mathcal{H}, Z \subseteq X$ and $x \in X$, $\Delta(h|_Z, x) \geq 0$.*

**Definition 7 (Adaptive Submodularity)** *A function $f : \mathcal{L}_{X,\mathcal{H}} \to \mathbb{R}_+$ is adaptive submodular if the conditional expected marginal benefit of a given item does not increase if the partial labeling is extended. That is, if for all $h \in \mathcal{H}$, for all $Z_1 \subseteq Z_2 \subseteq X$ ,and for all $x \in X$,*

$$\Delta(h|_{Z_1}, x) \geq \Delta(h|_{Z_2}, x).$$

The central theorem of adaptive submodularity, stated below as Theorem 9, links the expected utility of the optimal policy for maximizing $f_{\mathrm{avg}}$ with the expected utility of an approximately-greedy policy.

**Definition 8 (Approximate Greedy)** *Let $\alpha \geq 1$. A policy $\pi : \mathcal{L}_{X,\mathcal{H}} \to X$ is $\alpha$-approximately greedy with respect to a utility function $f$ if for every $h$ and for every $Z \subseteq X$*

$$\Delta(h|_Z, \pi(h|_Z)) \geq \frac{1}{\alpha} \max_{x \in X} \Delta(h|_Z, x). \tag{9}$$

**Theorem 9 (Golovin and Krause (2010))** *Let $f : \mathcal{L}_{X,\mathcal{H}} \to \mathbb{R}_+$ be a utility function, and let $\pi : \mathcal{L}_{X,\mathcal{H}} \to X$ be a policy. If $f$ is adaptive monotone and adaptive submodular, and $\pi$ is $\alpha$-approximately greedy, then for any policy $\pi^*$ and for all positive integers $n, k$,*

$$f_{\mathrm{avg}}(\pi, n) \geq (1 - e^{-\frac{n}{\alpha k}}) f_{\mathrm{avg}}(\pi^*, k). \tag{10}$$

In the active learning setting, we define the utility function $f$ as in Equation (7) and have the following result:

**Lemma 10 (Golovin and Krause (2010))** *The function $f$ defined in Equation (7) is adaptive monotone and adaptive submodular.*

Therefore Theorem 9 holds for this utility function. It follows that for any $\alpha$-approximate greedy policy $\pi$, any policy $\pi^*$ and any integers $n, k > 0$,

$$f_{\mathrm{avg}}(\pi, n) \geq (1 - e^{-\frac{n}{\alpha k}}) f_{\mathrm{avg}}(\pi^*, k) \geq f_{\mathrm{avg}}(\pi^*, k) - e^{-\frac{n}{\alpha k}}. \tag{11}$$

In order to prove that Equation (8) holds, we now show that the policy implemented by ALuMA is approximately greedy relative to $f$, with high probability over the randomization of ALuMA.

**Theorem 11** *With probability at least $1 - \delta/2$, the policy applied by ALuMA is a 4-approximately greedy policy with respect to the utility function $f$ defined in Equation (7).*

**Proof** Define $\lambda = 1/3$ and $\alpha = \left(\frac{1+\lambda}{1-\lambda}\right)^2 = 4$. We need to show that for any $Z \subseteq X$, our policy selects an element $x$ that approximately maximizes

$$\begin{aligned}
\Delta(h|_Z, x) &= \mathbb{E}_{g \sim U}\left[ f(g|_{Z \cup \{x\}}) - f(g|_Z) \mid g|_Z = h|_Z \right] \\
&= \Big( \mathbb{P}(V_{t,x}^1) \cdot (\mathbb{P}(V(h|_Z)) - \mathbb{P}(V(h|_Z \cup \{(x,1)\}))) + \\
&\qquad\qquad \mathbb{P}(V_{t,x}^{-1}) \cdot (\mathbb{P}(V(h|_Z) - \mathbb{P}(V(h|_Z \cup \{(x,-1)\})))) \Big) \\
&= 2 \, \mathbb{P}(V_{t,x}^1) \, \mathbb{P}(V_{t,x}^{-1}).
\end{aligned}$$

Therefore it suffices to show that for all $x \in X$ and for all iterations $t$,

$$\mathbb{P}(V_{t,x_t}^1) \, \mathbb{P}(V_{t,x_t}^{-1}) \geq \frac{1}{\alpha} \, \mathbb{P}(V_{t,x}^1) \, \mathbb{P}(V_{t,x}^{-1}).$$

where $x_t$ is the element chosen by ALuMA at iteration $t$.

By line 6 of Alg. 1, we select $x_t \in X_t$ that maximizes $\hat{v}_{x,1} \cdot \hat{v}_{x,-1}$, where

$$\hat{v}_{x,j} \leftarrow \text{VolEst}(V_{t,x}^j, \lambda, \frac{\delta}{4mN}) \quad , \quad j \in \{\pm 1\}.$$

Since ALuMA calls VolEst at most $2mN$ times in total, by Lemma 4 with probability $1 - \delta/2$, for all $x \in X, j \in \{\pm 1\}, t \in [N]$,

$$\frac{\mathbb{P}(V_{t,x}^j)}{1 + \lambda} < \hat{v}_{x,j} < \frac{\mathbb{P}(V_{t,x}^j)}{1 - \lambda}. \tag{12}$$

Therefore, for all $x \in X, j \in \{\pm 1\}, t \in [N]$,

$$\frac{\mathbb{P}(V_{t,x}^1) \, \mathbb{P}(V_{t,x}^{-1})}{(1 + \lambda)^2} < \hat{v}_{x,1} \cdot \hat{v}_{x,-1} < \frac{\mathbb{P}(V_{t,x}^1) \, \mathbb{P}(V_{t,x}^{-1})}{(1 - \lambda)^2}.$$

Let $x_t^* = \text{argmax}_{x \in X} \, \Delta(h|_X, x) = \text{argmax}_{x \in X} (\mathbb{P}(V_{t,x}^1) \, \mathbb{P}(V_{t,x}^{-1}))$. Then

$$\frac{\mathbb{P}(V_{t,x_t^*}^1) \, \mathbb{P}(V_{t,x_t^*}^{-1})}{(1 + \lambda)^2} < \hat{v}_{x_t^*,1} \cdot \hat{v}_{x_t^*,-1} < \hat{v}_{x_t,1} \cdot \hat{v}_{x_t,-1} < \frac{\mathbb{P}(V_{t,x_t}^1) \, \mathbb{P}(V_{t,x_t}^{-1})}{(1 - \lambda)^2}$$

Therefore

$$\mathbb{P}(V_{t,x_t}^1) \, \mathbb{P}(V_{t,x_t}^{-1}) \geq \frac{(1 - \lambda)^2}{(1 + \lambda)^2} \, \mathbb{P}(V_{t,x_t^*}^1) \, \mathbb{P}(V_{t,x_t^*}^{-1}).$$

Since $4 = \alpha = \frac{(1 + \lambda)^2}{(1 - \lambda)^2}$ this proves that our policy is 4-approximately greedy. ∎

## A.2 Comparing to OPT

By Theorem 11, ALuMA implements an approximately greedy policy. In addition, by Equation (11), any approximately greedy policy reduces the version space almost as fast as any other policy, including the optimal one. While these results seem promising, they are not enough for our needs. Recall that our true objective is not to have a small version space but rather to be able to correctly label all the examples in $X$. Therefore, we must quantify how the size of the version space corresponds to this objective.

The first issue with Equation (11) is that it does not provide a worst-case guarantee, since $f_{\text{avg}}$ averages over all $h \in \mathcal{H}$. It should be noted that Golovin and Krause (2010) derive a worst-case guarantee as well, but their approach requires that the utility function receive discrete values, which clearly does not hold in our case. The following lemma helps solve this issue, by providing a guarantee which holds individually for any $h \in H$.

**Lemma 12** *Let $\pi^*$ be a policy that achieves* OPT*, that is $c_{wc}(\pi^*) = $ OPT*. For any $h \in \mathcal{H}$, any $\alpha$-approximate greedy policy $\pi$, and any $n$,*

$$f_{\text{avg}}(\pi^*, \text{OPT}) - f_{\text{avg}}(\pi, n) \geq \mathbb{P}(V(h|_X)) \left( \mathbb{P}(V(S(\pi, h, n))) - \mathbb{P}(V(h|_X)) \right) \ .$$

**Proof** Since $\pi^*$ is an optimal policy, the version space induced by the labels $\pi^*$ queried within OPT iterations is exactly the set of hypotheses which are consistent with the true labels of the sample. Therefore, for any $h \in \mathcal{H}$.

$$\mathbb{P}(V(S(\pi^*, h, \text{OPT}))) = \mathbb{P}(V(h|_X)).$$

By definition of $f_{\text{avg}}$,

$$\begin{aligned}
f_{\text{avg}}(\pi^*, \text{OPT}) - f_{\text{avg}}(\pi, n) &= \mathbb{E}_{h \sim U}[\mathbb{P}(V(S(\pi, h, n))) - \mathbb{P}(V(S(\pi^*, h, \text{OPT})))] \\
&= \mathbb{E}_{h \sim U}[\mathbb{P}(V(S(\pi, h, n))) - \mathbb{P}(V(h|_X))].
\end{aligned}$$

Since $S(\pi, h, n)$ does not depend on the value of $h$ outside of $X$, we can sum over the possible labelings of $X$ to have

$$f_{\text{avg}}(\pi^*, \text{OPT}) - f_{\text{avg}}(\pi, n) = \sum_{h|_X : h \in \mathcal{H}} \mathbb{P}(V(h|_X))(\mathbb{P}(V(S(\pi, h|_X, n))) - \mathbb{P}(V(h|_X))).$$

Now, it is easy to see that for any $h \in \mathcal{H}$, $V(S(\pi, h|_X, n))) \supseteq V(h|_X)$, thus

$$\mathbb{P}(V(S(\pi, h|_X, n))) - \mathbb{P}(V(h|_X)) \geq 0.$$

It follows that for any $h \in \mathcal{H}$

$$f_{\text{avg}}(\pi^*, \text{OPT}) - f_{\text{avg}}(\pi, n) \geq \mathbb{P}(V(h|_X))(\mathbb{P}(V(S(\pi, h|_X, n))) - \mathbb{P}(V(h|_X))).$$

■

Combining Equation (11) and Lemma 12 we conclude that for any $\alpha$-approximate greedy policy $\pi$,

$$\forall h \in \mathcal{H}, \quad \mathbb{P}(V(h|_X))(\mathbb{P}(V(S(\pi, h|_X, n))) - \mathbb{P}(V(h|_X))) \ \leq \ e^{-\frac{n}{\alpha \text{OPT}}} \ ,$$

which yields

$$\forall h \in \mathcal{H}, \quad \frac{\mathbb{P}(V(h|_X))}{\mathbb{P}(V(S(\pi, h, n)))} \geq \frac{\mathbb{P}(V(h|_X))^2}{e^{-\frac{n}{\alpha \text{OPT}}} + \mathbb{P}(V(h|_X))^2}. \tag{13}$$

This means that if $\mathbb{P}(V(h|_X))$ is large enough and we run an approximate greedy policy, then after a sufficient number of iterations, most of the remaining version space induces the correct labeling of the sample. We now show that under the margin assumption, $\mathbb{P}(V(h|_X))$ is indeed bounded from below.

### A.3 Using the Margin Assumption

Denote by $\mathcal{H}_\gamma$ the subset of $\mathcal{H}$ which has margin $\gamma$ on the unlabeled sample $X$, that is

$$\mathcal{H}_\gamma = \{h \in \mathcal{H} \mid \exists w \in \mathbb{B}_1^d \text{ s.t. } \forall x \in X, \, h(x)\langle w, x \rangle \geq \gamma\}.$$

Our bound on the number of queries for a hypothesis with a margin hinges on the fact that $\mathbb{P}(V(h|_X)$ is large enough if $h$ has a margin on $X$. This is quantified in the following lemma.

**Lemma 13** *For all $h \in \mathcal{H}_\gamma$, $\mathbb{P}(V(h|_X)) \geq \left(\frac{\gamma}{2}\right)^d$.*

**Proof** Fix some $h \in \mathcal{H}_\gamma$. Let $w \in \mathbb{B}_1^d$ such that $\forall x \in X, \, h(x)\langle w, x \rangle \geq \gamma$. For a given $v \in \mathbb{B}_1^d$, denote by $h_v \in \mathcal{H}$ the mapping $x \mapsto \text{sgn}(\langle v, x \rangle)$. Note that for all $v \in \mathbb{B}_1^d$ such that $\|w - v\| < \gamma$, $h_v \in V(h|_X)$. This is because for all $x \in X$,

$$h(x)\langle v, x \rangle = \langle v - w, h(x) \cdot x \rangle + h(x)\langle w, x \rangle$$
$$\geq -\|w - v\| \cdot \|h(x) \cdot x\| + \gamma > -\gamma + \gamma = 0,$$

which implies $\text{sgn}(\langle v, x \rangle) = h(x)$. It follows that $\{v \mid h_v \in V(h|_X)\} \supseteq \mathbb{B}_1^d \cap B(w, \gamma)$, where $B(z, r)$ denotes the ball of radius $r$ with center at $z$. Let $u = (1 - \gamma/2)w$. Then for any $z \in B(u, \gamma/2)$, we have $z \in \mathbb{B}_1^d$, since

$$\|z\| = \|z - u + u\| \leq \|z - u\| + \|u\| \leq \gamma/2 + 1 - \gamma/2 = 1.$$

In addition, $z \in B(w, \gamma)$ since

$$\|z - w\| = \|z - u + u - w\| \leq \|z - u\| + \|u - w\| \leq \gamma/2 + \gamma/2 = \gamma.$$

Therefore $B(u, \gamma/2) \subseteq \mathbb{B}_1^d \cap B(w, \gamma)$. We conclude that $\{v \mid h_v \in V(h|_X)\} \supseteq B(u, \gamma/2)$. Thus,

$$\mathbb{P}(V(h|_X)) \geq \text{Vol}(B(u, \gamma/2))/\text{Vol}(\mathbb{B}_1^d) \geq \left(\frac{\gamma}{2}\right)^d.$$

■

We can now ensure that most of the remaining version space induces the correct labeling. The following theorem is a direct consequence of combining Lemma 13 with Equation (13).

**Theorem 14** *For any $h \in \mathcal{H}_\gamma$, any $\alpha$-approximate greedy policy $\pi$, and any*

$$n \geq \alpha \cdot \text{OPT} \cdot (2d \ln(2/\gamma) + \ln(2)),$$

*we have*

$$\frac{\mathbb{P}(V(h|_X))}{\mathbb{P}(V(S(\pi, h, n)))} > \frac{2}{3}. \tag{14}$$

We are now ready to prove items (1) and (2) of Theorem 2.

**Corollary 15** *If ALuMA is executed for $n \geq 4 \cdot \mathrm{OPT} \cdot (2d \ln(2/\gamma) + \ln(2))$ iterations, then with probability at least $1 - \delta$ ALuMA returns the correct labeling of all the elements of $X$.*

**Proof** By Theorem 11, with probability at least $1 - \delta/2$ ALuMA runs a 4-approximately greedy policy. Therefore, if $n$ satisfies our assumption, then Equation (14) holds. Since $V(h|_X) \subseteq V(S(\pi, h, n)) = V_n$, it follows that the probability of drawing a hypothesis from $V(h|_X)$ when drawing uniformly from $V_n$ is at least $\frac{2}{3}$. In step 12 of ALuMA, $M \geq 72 \ln(2/\delta)$ hypotheses are drawn $\frac{1}{12}$-uniformly at random from $V_n$. Therefore each hypothesis $h_i$ is from $V(h|_X)$ with probability at least $\frac{7}{12}$. By Hoeffding's inequality,

$$\mathbb{P}(\frac{1}{M} \sum_{i=1}^{M} I[h_i \in V(h|_X)] \leq \frac{1}{2}] \leq \exp(-M/72) = \frac{\delta}{2}.$$

Therefore, with probability $1 - \delta/2$ the majority vote over the drawn hypotheses provides the correct label for all $x \in X$. In total, the correct labeling is returned with probability $1 - \delta$. ■

## Appendix B. Analysis of the Preprocessing Procedure

We now prove Theorem 3 by showing that Alg. 2 satisfies the claims of the theorem. It is clear that Alg. 2 is polynomial as required in item (3). In addition, item (1) holds from the definition of Alg. 2. We have left to prove item (2). We first prove that it holds for the case where the input is represented directly as $X \subseteq \mathbb{R}^d$.

We start by showing that under the assumption of Theorem 3, the set $\{x'_1, \ldots, x'_m\}$, which is generated in step 8, is separated with a bounded margin by the original labels of $x_i$. Fix $\gamma > 0$ and $w^* \in \mathbb{B}_1^d$. For each $i \in [m]$, define

$$\ell_i = \max(0, \gamma - L(i)\langle w^*, x_i \rangle).$$

Thus, $\ell_i$ quantifies the margin violation of example $x_i$ by $w^*$, relative to its true label $L(i)$.

**Lemma 16** *If $H \geq \sum_{i=1}^{m} \ell_i^2$, where $H$ is the input to Alg. 2, then there is a $w \in \mathbb{B}_1^{d+m}$ such that for all $i \in [m]$, $L(i)\langle w, x'_i \rangle \geq \frac{\gamma}{1+\sqrt{H}}$.*

**Proof** By step 8 in Alg. 2, $x'_i = (a \cdot x_i; \sqrt{1-a^2} \cdot e_i)$, where $a = \sqrt{\frac{1}{1+\sqrt{H}}}$. Define

$$w' = (w^*; \frac{a}{\sqrt{1-a^2}}(L(1)\ell_1, \ldots, L(m)\ell_m)).$$

Then

$$L(i)\langle w', x'_i \rangle = aL(i)\langle w^*, x_i \rangle + a\ell_i \geq a(\gamma - \ell_i) + a\ell_i = a\gamma.$$

Let $w = \frac{w'}{\|w'\|}$. Then $w \in \mathbb{B}_1^{d+m}$, and

$$L(i)\langle w, x'_i \rangle = \frac{L(i)\langle w', x'_i \rangle}{\|w'\|} \geq \frac{a\gamma}{\sqrt{1 + \frac{a^2}{1-a^2}\sum_{i=1}^{m}\ell_i^2}} = \frac{\gamma}{\sqrt{\frac{1}{a^2} + \frac{1}{1-a^2}\sum_{i=1}^{m}\ell_i^2}}.$$

Set $a^2 = \frac{1}{1+\sqrt{H}}$, and assume $H \geq \sum_{i=1}^m \ell_i^2$. Then

$$L(i)\langle w, x_i' \rangle \geq \frac{\gamma}{1+\sqrt{H}}.$$

$\blacksquare$

The set $\{\bar{x}_1, \ldots, \bar{x}_m\}$ returned by Alg. 2 is a Johnson-Lindenstrauss projection of $\{x_1', \ldots, x_m'\}$ on $\mathbb{R}^k$. It is known (see e.g. Balcan et al. (2006b)) that if a set of $m$ points is separable with margin $\eta$ and $k \geq O\left(\frac{\ln(m/\delta)}{\eta^2}\right)$, then with probability $1 - \delta$, the projected points are separable with margin $\eta/2$. Setting $\eta = \frac{\gamma}{1+\sqrt{H}}$, it is easy to see that step 12 in Alg. 2 indeed maintains the desired margin. This completes the proof of item (2) of Theorem 3 for the case where the input is $X \subseteq \mathbb{R}^m$.

We now show that if the input is a kernel matrix $K$, then the decomposition step 3 preserves the separation properties of the input data, thus showing that item (2) holds in this case as well. To show that our decomposition step does not change the properties of the original data, we first use the following lemma, which indicates that separation properties are conserved under different decompositions of the same kernel matrix.

**Lemma 17 (Sabato et al. (2010), Lemma 6.3)** *Let $K \in \mathbb{R}^{m \times m}$ be an invertible PSD matrix and let $V \in \mathbb{R}^{m \times n}, U \in \mathbb{R}^{m \times k}$ be matrices such that $K = VV^T = UU^T$. For any vector $w \in \mathbb{R}^n$ there exists a vector $u \in \mathbb{R}^k$ such that $Vw = Uu$ and $\|u\| \leq \|w\|$.*

The next lemma extends the above result, showing that the property holds even if $K$ is not invertible.

**Lemma 18** *Let $K \in \mathbb{R}^{m \times m}$ be a PSD matrix and let $V \in \mathbb{R}^{m \times n}, U \in \mathbb{R}^{m \times k}$ be matrices such that $K = VV^T = UU^T$. For any vector $w \in \mathbb{R}^n$ there exists a vector $u \in \mathbb{R}^k$ such that $Vw = Uu$ and $\|u\| \leq \|w\|$.*

**Proof** For a matrix $A$ and sets of indexes $I, J$ let $A[I]$ be the sub-matrix of $A$ whose rows are the rows of $A$ with an index in $I$. Let $A[I, I]$ be the sub-matrix of $A$ whose rows and columns are those that have index $I$ in $A$.

If $K$ is invertible, the claim holds by Lemma 17. Thus, assume $K$ is singular. Let $I \subseteq [m]$ be a maximal subset such that the matrix $K[I; I]$ is invertible—If no such subset exists then $K, V, U$ are all zero and the claim is trivial. By Lemma 17, $K[I; I] = V[I](V[I])^T = U[I](U[I])^T$, and there exists a vector $u$ such that $V[I]w = U[I]u$, and $\|u\| \leq \|w\|$. We will show that for any $i \notin I$, $V[i]w = U[i]u$ as well.

For any $i \notin I$, $K[I \cup \{i\}; I \cup \{i\}]$ is singular. Therefore $V[I \cup \{i\}]$ is singular, while $V[I]$ is not. Thus there is some vector $\lambda \in \mathbb{R}^{|I|}$ such that $V[i]^T = V[I]^T \lambda$. By a similar argument there is some vector $\eta \in \mathbb{R}^{|I|}$ such that $U[i]^T = U[I]^T \eta$. We have $K[I, i] = V[I]V[i]^T = V[I]V[I]^T \lambda = K[I, I]\lambda$. Similarly for $U$, $K[I, i] = K[I, I]\eta$. Therefore $K[I, I](\lambda - \eta) = 0$. Since $K[I, I]$ is invertible, it follows that $\lambda = \eta$. Therefore, $U[i]u = \eta^T U[I]u = \lambda^T V[I]w = V[i]w$. $\blacksquare$

We now use this lemma to show that the decomposition step does not change the upper bound on the margin loss which is assumed in Theorem 3.

**Theorem 19** *Let $\psi_1, \ldots, \psi_m$ be a set of vectors in a Hilbert space $S$, and let $K \in \mathbb{R}^{m \times m}$ such that for all $i, j \in [m]$, $K_{i,j} = \langle \psi_i, \psi_j \rangle$. suppose there exists a $w \in S$ with $\|w\| \leq 1$ such that*

$$H \geq \sum_{i=1}^{m} \max(0, \gamma - y_i \langle w, \psi_i \rangle)^2. \tag{15}$$

*Let $U \in \mathbb{R}^{m \times k}$ such that $K = UU^T$ and let $x_i$ be row $i$ of $U$. Then there exists a $u \in \mathbb{B}_1^k$ such that*

$$H \geq \sum_{i=1}^{m} \max(0, \gamma - y_i \langle u, x_i \rangle)^2. \tag{16}$$

**Proof** Let $\alpha_1, \ldots, \alpha_n \in S$ be an orthogonal basis for the span of $\psi_1, \ldots, \psi_m$ and $w$, and let $v_1, \ldots, v_m, v_w \in \mathbb{R}^n$ such that $\sum_{l=1}^{n} v_i[l]\alpha_l = \psi_i$ and $\sum_{l=1}^{n} v_w[l]\alpha_l = w$. Let $V \in \mathbb{R}^{m \times n}$ be a matrix such that row $i$ of the matrix is $v_i$. Then $K = VV^T$, and $Vv_w = r$, where $r[i] = \langle v_w, v_i \rangle = \langle w, \psi_i \rangle$. By Lemma 18, there exists a $u \in \mathbb{R}^k$ such that $Uu = r$. Then we have $\langle u, x_i \rangle = r[i]$. Therefore for all $i \in [m]$, $\langle w, \psi_i \rangle = \langle u, x_i \rangle$, thus Equation (15) implies Equation (16). In addition, $\|u\| \leq \|v_w\| = \|w\| \leq 1$, therefore $u \in \mathbb{B}_1^k$. ∎

## Appendix C. Analysis of the Simpler Implementation

The following theorem shows that using the estimation procedure listed in Alg. 4 also results in an approximately greedy policy, as does the original implementation of ALuMA. Similarly to the proof of Corollary 15, it follows that Theorem 2 holds for this implementation as well.

**Theorem 20** *If for each iteration $t$ of the algorithm, the greedy choice $x^*$ satisfies*

$$\forall j \in \{-1, +1\}, \quad \mathbb{P}[h(x^*) = j \mid h \in V_t] \geq 4\sqrt{\lambda}$$

*then ALuMA with the estimation procedure implements a 2-approximate greedy policy. Moreover, it is possible to efficiently verify that this condition holds while running the algorithm.*

**Proof** Fix the iteration $t$, and denote $p_{x,1} = \mathbb{P}(V_{t,x}^1)/\mathbb{P}(V_t)$ and $p_{x,-1} = \mathbb{P}(V_{t,x}^1)/\mathbb{P}(V_t)$. Note that $p_{x,1} + p_{x,-1} = 1$. Since $h_1, \ldots, h_k$ are sampled $\lambda$-uniformly from the version space, we have

$$\forall i \in [k], |\mathbb{P}[h_i \in V_{t,x}^j] - p_{x,j}| \leq \lambda. \tag{17}$$

In addition, by Hoeffding's inequality and a union bound over the examples in the pool and the iterations of the algorithm,

$$\mathbb{P}[\exists x, |\hat{v}_{x_i,j} - \mathbb{P}[h_i \in V_{t,x}^j]| \geq \lambda] \leq 2m \exp(-2k\lambda^2). \tag{18}$$

From Alg. 4 we have $k = \frac{\ln(2m/\delta)}{2\lambda^2}$. Combining this with Equation (17) and Equation (18) we get that

$$\mathbb{P}[\exists x, |\hat{v}_{x_i,j} - p_{x_i,j}| \geq 2\lambda] \leq \delta.$$

The greedy choice for this iteration is

$$x^* \in \operatorname*{argmax}_{x \in X} \Delta(h|_X, x) = \operatorname*{argmax}_{x \in X} (\mathbb{P}(p_{x,1}) \, \mathbb{P}(p_{x,-1})).$$

By the assumption in the theorem, $p_{x^*,j} \geq 4\sqrt{\lambda}$ for $j \in \{-1, +1\}$. Since $\lambda \in (0, \frac{1}{64})$, we have $\lambda \leq \sqrt{\lambda}/8$. Therefore $p_{x^*,j} - 2\lambda \geq 4\sqrt{\lambda} - \sqrt{\lambda}/4 \geq \sqrt{10\lambda}$. Therefore

$$\hat{v}_{x^*,1} \hat{v}_{x^*,-1} \geq (p_{x^*,1} - 2\lambda)(p_{x^*,-1} - 2\lambda) \geq 10\lambda. \tag{19}$$

Let $\tilde{x} = \operatorname{argmax}(\hat{v}_{x,-1} \hat{v}_{x,+1})$ be the query selected by ALuMA using Alg. 4. Then

$$\hat{v}_{x^*,-1} \hat{v}_{x^*,+1} \leq \hat{v}_{\tilde{x},-1} \hat{v}_{\tilde{x},+1} \leq (p_{\tilde{x},1} + 2\lambda)(p_{\tilde{x},-1} + 2\lambda) \leq p_{\tilde{x},1} p_{\tilde{x},-1} + 4\lambda.$$

Where in the last inequality we used the facts that $p_{\tilde{x},1} + p_{\tilde{x},-1} = 1$ and $4\lambda^2 \leq 2\lambda$. On the other hand, by Equation (19)

$$\hat{v}_{x^*,-1} \hat{v}_{x^*,+1} \geq 5\lambda + \frac{1}{2} \hat{v}_{x^*,-1} \hat{v}_{x^*,+1} \geq 5\lambda + \frac{1}{2}(p_{x^*,-1} - 2\lambda)(p_{x^*,-1} - 2\lambda) \geq 4\lambda + \frac{1}{2} p_{x^*,-1} p_{x^*,-1}.$$

Combining the two inequalities for $\hat{v}_{x^*,-1} \hat{v}_{x^*,+1}$ it follows that $p_{\tilde{x},1} p_{\tilde{x},-1} \geq \frac{1}{2} p_{x^*,-1} p_{x^*,-1}$, Thus this is a 2-approximately greedy policy.

To verify that the assumption holds at each iteration of the algorithm, note that for all $x = x_i$ such that $i \in I_t$

$$p_{x,-1} p_{x,+1} \geq (\hat{v}_{x,-1} - 2\lambda)(\hat{v}_{x,+1} - 2\lambda) \geq \hat{v}_{x,-1} \hat{v}_{x,+1} - 2\lambda.$$

therefore it suffices to check that for all $x = x_i$ such that $i \in I_t$

$$\hat{v}_{x,-1} \hat{v}_{x,+1} \geq 4\sqrt{\lambda} + 2\lambda.$$

∎